%% file: colm2024_conference.tex
\documentclass{article} %
\usepackage{colm2024_conference}

\usepackage{booktabs}
\usepackage{graphicx}
\usepackage{enumitem}
\usepackage{wrapfig}
\usepackage{algorithm}
\usepackage{algpseudocode}
\usepackage{threeparttable}
\usepackage{CJKutf8}
\usepackage{chngcntr} 

\usepackage[raster,skins]{tcolorbox} %
\tcbuselibrary{listings,skins,breakable}

\usepackage{xcolor}
\usepackage{listings}

\usepackage{microtype}
\usepackage{amsmath}
\usepackage{amsfonts}
\usepackage{colortbl}
\usepackage[utf8]{inputenc}
\usepackage[T1]{fontenc}
\definecolor{lightgray}{rgb}{0.9,0.9,0.9}
\usepackage{caption}
\usepackage{subcaption}
\usepackage{xcolor}
\usepackage{setspace}
\usepackage{url}
\usepackage{multirow}
\usepackage{colortbl}
\usepackage{tabularx}
\usepackage{blindtext}
\usepackage{pgfplots}
\pgfplotsset{compat=1.18} 
\usepackage{tikz}
\usetikzlibrary{er,positioning,bayesnet}
\usepackage{makecell}
\usepackage{tipa}
\usepackage{siunitx}
\usepackage{nicefrac}
\usepackage{tocloft}
\usepackage{listings}
\usepackage{xltabular}
\usepackage{adjustbox}
\usepackage{xurl}
\usepackage{xcolor}
\usepackage{cleveref}

\input{math_commands.tex}

\title{Qwen3-ASR Technical Report}

\author{
\bf Qwen Team
}

\definecolor{lightgraybg}{RGB}{240, 240, 240}

\begin{document}

\maketitle

\begin{abstract}

In this report, we introduce Qwen3-ASR family, which includes two powerful all-in-one speech recognition models and a novel non-autoregressive speech forced alignment model. \textit{Qwen3-ASR-1.7B} and \textit{Qwen3-ASR-0.6B} are ASR models that support language identification and ASR for 52 languages and dialects. Both of them leverage large-scale speech training data and the strong audio understanding ability of their foundation model Qwen3-Omni. We conduct comprehensive internal evaluation besides the open-sourced benchmarks as ASR models might differ little on open-sourced benchmark scores but exhibit significant quality differences in real-world scenarios. The experiments reveal that the 1.7B version achieves state-of-the-art performance among open-sourced ASR models and is competitive with the strongest proprietary APIs while the 0.6B version offers the best accuracy–efficiency trade-off. Qwen3-ASR-0.6B can achieve an average time-to-first-token as low as 92ms and transcribe 2,000 seconds speech in 1 second at a concurrency of 128. \textit{Qwen3-ForcedAligner-0.6B} is an LLM based NAR timestamp predictor that is able to align text-speech pairs in 11 languages. Timestamp accuracy experiments show that the proposed model outperforms the three strongest force alignment models and takes more advantages in efficiency and versatility. To further accelerate the community research of ASR and audio understanding, we release these models under the Apache 2.0 license.

\end{abstract}

\input{content/intro.tex}
\input{content/method}

\input{content/experiments.tex}
\input{content/conclusion.tex}
\input{content/authors.tex}

\bibliography{biblio}
\bibliographystyle{colm2024_conference}

\input{content/appendix}

\end{document}

%% file: math_commands.tex
\usepackage{amsmath,amsfonts,bm}

\def\eqref#1{equation~\ref{#1}}

\def\1{\bm{1}}

\DeclareMathAlphabet{\mathsfit}{\encodingdefault}{\sfdefault}{m}{sl}
\SetMathAlphabet{\mathsfit}{bold}{\encodingdefault}{\sfdefault}{bx}{n}



%% file: content/intro.tex
\section{Introduction}
\label{sec:intro}

\begin{figure}[ht]
    \centering
    \includegraphics[width=\linewidth]{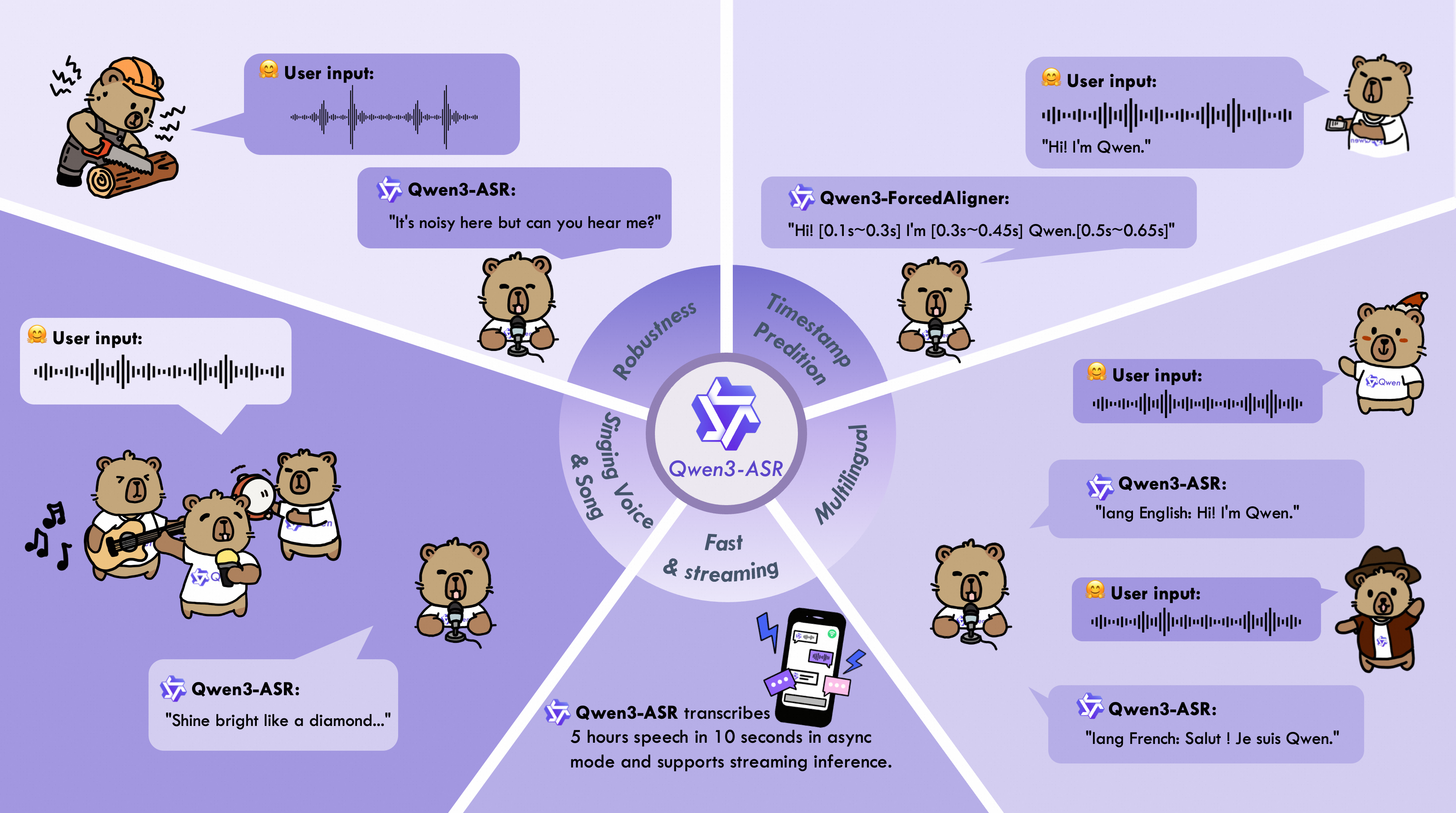}
    \caption{Qwen3-ASR family includes all-in-one ASR models with advantages in multilingual, noisy speech recognition, singing voice recognition and inference efficiency, so as to a novel multilingual speech forced alignment model for predicting timestamps of words or sentences in ASR results.}
    \label{fig:placeholder}
\end{figure}

In recent years, Automatic Speech Recognition~(ASR) has transitioned from traditional end-to-end~(E2E) paradigms, e.g., Transducer~\citep{arxiv2012-alex-rnnt} and AED~\citep{icassp2016-William-las,icml2023-alec-Whisper} to the Large Audio-Language Model~(LALM) paradigm. Compared with traditional ASR, this paradigm can take advantage of the language modeling capabilities and world knowledge of large language models. The model first forms a high-level understanding of the audio signal and then generates transcription conditioned on this understanding, rather than relying solely on bottom-up acoustic pattern matching.
Under this paradigm, issues that are relatively challenging for conventional ASR models—such as long-form transcription, robustness to noise, world-knowledge and named-entity recognition, as well as multilingual and dialectal coverage, can be addressed more naturally.

In real-world deployments, ASR systems are often required to output timestamps alongside transcripts (e.g., for subtitle generation). Prior work typically performs timestamping as a post-processing step using techniques such as CTC or CIF~\citep{Ludwig2020ctc, Elena2023nfa, shi2023achieving}. We would like to highlight that an LALM-based approach can yield more accurate and faster timestamp prediction at arbitrary temporal granularities, and that, by leveraging the multilingual capacity of LALMs, a single unified model can provide timestamp alignment across diverse languages.



In this report, we present the Qwen3-ASR family, including \textit{Qwen3-ASR-1.7B} and \textit{Qwen3-ASR-0.6B} - two all-in-one ASR models with language identification~(LID) ability for 52 languages and dialects, and \textit{Qwen3-ForcedAligner-0.6B} - the first lightweight LALM-based multilingual forced aligner supporting 11 languages and flexible timestamp prediction granularities. These models are posttrained from the strong foundation model of Qwen3-Omni~\citep{qwen3-omni}. For evaluating the performance of ASR models on benchmarks out of the open-sourced ones~(ASR models at present have reached the limit of annotation errors on several test sets), we build a series of internal benchmarks covering more than complex acoustic environment, dialects, elders and kids speech and multilingual. Qwen3-ASR-1.7B achieves state-of-the-art~(SOTA) performance among open-sourced ASR models and is competitive with the strongest proprietary commercial APIs. Qwen3-ASR-0.6B offers the best accuracy-model-size trade-off, making it a strong choice for on-device deployment. Qwen3-ForcedAligner-0.6B delivers highly accurate forced-alignment timestamps and inherits the key capabilities of Qwen3-ASR, including multilingual and long-form speech support, enabling scalable labeling of speech-transcript pairs.

The key features and contributions of the proposed Qwen3-ASR family models can be summarized as:

\begin{itemize}

\item \textbf{Achieves state-of-the-art all-in-one ASR and LID performance.} Qwen3-ASR-1.7B and Qwen3-ASR-0.6B finely support 30 languages, 22 Chinese dialects ASR, and English from countries and regions worldwide. These two models also conduct robust speech recognition under complex environment, including but not limited to singing voice and song recognition, noise environment recognition and complex text patterns recognition.

\item \textbf{Presents novel speech force alignment architecture.} To the best of our knowledge, we introduce the first Large Language Model based speech forced aligner that produces accurate timestamps at flexible granularities, including word, sentence, and paragraph levels. In contrast to existing tools such as the Montreal Forced Aligner (MFA) and NeMo Forced Aligner (NFA), our model, Qwen3-ForcedAligner-0.6B, offers a unified, multilingual solution that addresses the lack of an all-in-one forced alignment system within the Qwen3-ASR family and fulfills a critical functional component for comprehensive spoken language processing.

\item \textbf{Open-source models and a comprehensive inference and fine-tuning framework.} In addition to releasing the weights of three models, we provide a fully open-source, user-friendly codebase that supports inference with multiple features (e.g., multi-granularity alignment, streaming transcription, and multilingual processing) as well as a reproducible fine-tuning recipe. We hope this unified toolkit will accelerate research and development efforts in the automatic speech recognition community.

\end{itemize}

%% file: content/method.tex
\newcommand{\prompt}[1]{\lstinline[style=promptstyle]!#1!}

\section{Qwen3-ASR}

\subsection{Architecture}

\begin{figure}[h]
    \centering
    \includegraphics[width=1.0\linewidth]{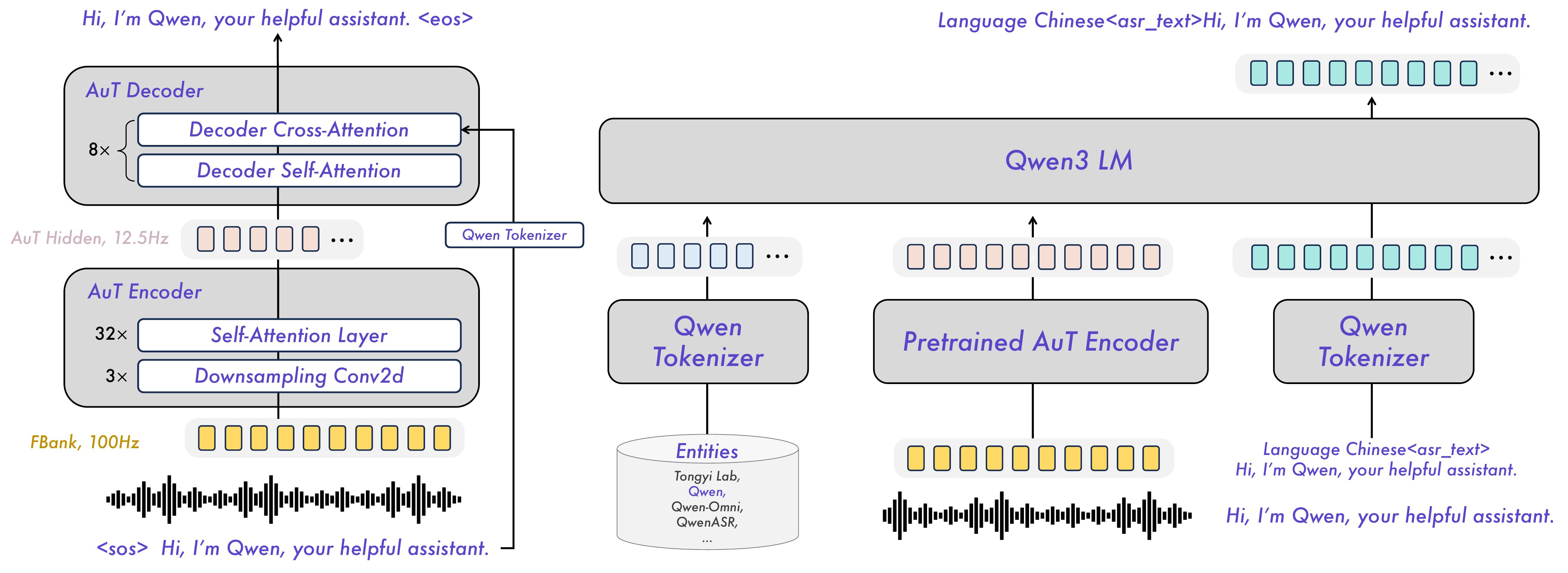}
    \caption{Architecture of AuT~(left) and the overview of Qwen3-ASR~(right).}
    \label{fig:aut}
\end{figure}

Qwen3-ASR family models leverage Qwen3-Omni as a foundation model, which proved to obtain strong audio understanding ability~\citep{arxiv2025-jinxu-qwen3omni}. Speech to recognize is first fed to AuT encoder, which is pretrained separately from Qwen3-Omni and Qwen3-ASR. As illustrated in Figure~\ref{fig:aut}(left), AuT is an attention-encoder-decoder~(AED) based ASR model which conducts 8 times downsampling to Fbank feature with 128 dimensions, yielding a 12.5Hz token rate audio encoder. We use a dynamic flash attention window size ranging from 1s to 8s, which allows Qwen3-ASR to perform both streaming inference with short chunks and offline inference with long queries. The architecture of models we are releasing is illustrated as Figure~\ref{fig:aut}(right) and detailed as below: \textbf{Qwen3-ASR-1.7B} is built with Qwen3-1.7B, a projector, and an AuT encoder with 300M parameters and 1024 hidden size. This model demonstrates strong performance in multilingual and dialect speech recognition, so as well as robustness under complex acoustic environment and text patterns. \textbf{Qwen3-ASR-0.6B} is built with Qwen3-0.6B, a projector and an AuT encoder with 180M parameters and a hidden size of 896. We design this compact model to balance recognition accuracy and inference efficiency, while remaining highly competitive among sub-1B-parameter ASR models.



\subsection{Training Strategies}
\label{section:traning-strategies}

The training process of Qwen3-ASR consists of AuT pretraining, Omni pretraining, and ASR post-training, where the first two stages are identical to those of Qwen3-Omni.
\begin{enumerate}[label=(\arabic*)]
    \item \textbf{AuT pretraining.} In this stage, we aim to obtain a pretrained encoder under the AED framework using large-scale labeled data. We leverage approximately 40 million hours of pseudo-labeled ASR data, where the majority is in Chinese and English. This pretrained encoder is shown to provide general and stable audio representations under dynamic attention window sizes.

    \item \textbf{Omni pretraining.} We use the pretrained Qwen3-Omni model as the foundation model for ASR training. Omni pretraining is conducted on multi-task audio, vision, and text data. In this stage, both Qwen3-ASR-0.6B and Qwen3-ASR-1.7B are trained with 3 trillion tokens, acquiring multi-modal understanding capability. The detailed training pipeline follows \citep{arxiv2025-jinxu-qwen3omni}.

    \item \textbf{ASR supervised finetuning (SFT).} In the SFT stage, we perform style transfer on the ASR input/output format with a substantially smaller set of multilingual data that is disjoint from the pretraining corpus. Besides standard Chinese, English and multilingual ASR data, SFT stage also utilizes non-speech data, streaming-enhancement data and context biasing data. Specifically, we train the model to be an ASR-only model that does not follow natural-language instructions in the prompt, in order to mitigate instruction injection and instruction-following failures. The output of Qwen3-ASR has outputs in two types for the given audio, with and without recognizable human speech:

    \definecolor{chatmlblue}{RGB}{0, 0, 255}
    \begin{CJK*}{UTF8}{gkai}
    \resizebox{0.925\textwidth}{!}{%
    \begin{tcolorbox}[
        colback=lightgray,
        colframe=black,
        boxrule=1pt,
        width=\textwidth,
        title={Qwen3-ASR output style},
        colbacktitle=black,
        coltitle=white,
        fonttitle=\bfseries\sffamily,
        halign title=left,
    ]
    \textbf{Output style 1: For recognizable speech}\\
    \textbf{<|im\_start|>}assistant \\
    language English\textbf{<asr\_text>}Today we release models including Qwen3-ASR-1.7B.\textbf{<|im\_end|>} \\
    
    \textbf{Output style 2: For no speech detected}\\
    \textbf{<|im\_start|>}assistant \\
    language None\textbf{<asr\_text><|im\_end|>}
    \end{tcolorbox}
    }
    \end{CJK*}
    Meanwhile, the model learns to utilize the context tokens inside the system prompt as background knowledge, allowing users to obtain customized ASR results.
    

    

    \item \textbf{ASR reinforcement learning (RL).} At the last stage, we use Group Sequence Policy Optimization~(GSPO, ~\cite{zheng2025group}) for further improving the quality of recognition. It turns out that RL plays an essential role in models' noise robustness, transcription stability and ability to analyze difficult cases. The total data leveraged by RL stage is about 50k utterances including 35\% Chinese and English data, 35\% multilingual data and 30\% functional data, which aims at improving transcribing stability in complex environments.
\end{enumerate}

\subsection{Features}
\label{method:features}

With the architecture and training strategies introduced above, Qwen3-ASR family models are notable in the aspects below:

\begin{table}[ht]
\centering
\caption{\textbf{Features of the Qwen3-ASR model family. Qwen3-ASR-1.7B and Qwen3-ASR-0.6B support 52 languages and dialects, comprising 30 languages and 22 Chinese dialects. Qwen3-ForcedAligner-0.6B supports 11 languages. Seq. Len. denotes the maximum audio length for single inference in seconds, and NAR denotes non-autoregressive inference.}}
\label{tab:feature-new}
\renewcommand{\arraystretch}{1.15}
\setlength{\tabcolsep}{2.5pt}
\footnotesize

\resizebox{\textwidth}{!}{%
\begin{tabular}{
>{\centering\arraybackslash}m{3.2cm}
>{\raggedright\arraybackslash}m{4.8cm}
>{\raggedright\arraybackslash}m{5.4cm}
>{\centering\arraybackslash}m{2.2cm}
>{\centering\arraybackslash}m{1.8cm}
>{\centering\arraybackslash}m{2.8cm}
}
\toprule
\multirow{2}{3.2cm}{\centering \textbf{Model}} & \multirow{2}{4.8cm}{\raggedright \textbf{Supported Languages}} & \multirow{2}{5.4cm}{\raggedright \textbf{Supported Dialects}} & \multirow{2}{2.2cm}{\centering \textbf{Inference Mode}} & \multirow{2}{1.8cm}{\centering \textbf{Seq. Len.}} & \multirow{2}{2.8cm}{\centering \textbf{Audio Types}} \\
& & & & & \\
\midrule

\multirow{8}{3.2cm}{\centering \textbf{Qwen3-ASR-1.7B} \&\\ \textbf{Qwen3-ASR-0.6B}}
&
\makecell[l]{Chinese (zh), English (en),\\
Cantonese (yue), Arabic (ar),\\
German (de), French (fr),\\
Spanish (es), Portuguese (pt),\\
Indonesian (id), Italian (it),\\
Korean (ko), Russian (ru),\\
Thai (th), Vietnamese (vi),\\
Japanese (ja), Turkish (tr),\\
Hindi (hi), Malay (ms),\\
Dutch (nl), Swedish (sv),\\
Danish (da), Finnish (fi),\\
Polish (pl), Czech (cs),\\
Filipino (fil), Persian (fa),\\
Greek (el), Hungarian (hu),\\
Macedonian (mk), Romanian (ro)}
&
\makecell[l]{
Anhui, Dongbei, Fujian, Gansu,\\
Guizhou, Hebei, Henan, Hubei,\\
Hunan, Jiangxi, Ningxia,\\
Shandong, Shaanxi, Shanxi,\\
Sichuan, Tianjin, Yunnan, Zhejiang.\\
Cantonese (Hong Kong accent),\\
Cantonese (Guangdong accent),\\
Wu language, Minnan language.}
&
\multirow{8}{2.2cm}{\centering Offline / \\ Streaming}
&
\multirow{8}{1.8cm}{\centering 1200s}
&
\multirow{8}{2.8cm}{\centering Speech,\\ Singing Voice,\\ Songs with BGM}
\\
\midrule

\multirow{3}{3.2cm}{\centering \textbf{Qwen3-\\ ForcedAligner-0.6B}}
&
\makecell[l]{Chinese, English, Cantonese,\\
French, German, Italian,\\
Japanese, Korean, Portuguese,\\
Russian, Spanish}
&
\multirow{3}{5.4cm}{\centering --}
&
\multirow{3}{2.2cm}{\centering NAR}
&
\multirow{3}{1.8cm}{\centering 300s}
&
\multirow{3}{2.8cm}{\centering Speech}
\\
\bottomrule
\end{tabular}%
}
\end{table}

\begin{enumerate}[label=(\arabic*)]
    \item \textbf{Accurate Chinese and English ASR.} Chinese and English account for the majority of the training data across all stages, and the model achieves leading Chinese and English recognition performance over multiple benchmarks compared with many competing systems.
    \item \textbf{Multilingual, multiple dialects supporting.} Qwen3-ASR-1.7B and Qwen3-ASR-0.6B support 30 languages and 22 dialects, detailed in~\Cref{tab:feature-new}.
    \item \textbf{Long-form and streaming inference.} Qwen3-ASR-1.7B and Qwen3-ASR-0.6B naturally support single speech no longer than 20 minutes and streaming/offline unified inference.
    \item \textbf{Singing voice and songs recognition.} Qwen3-ASR-1.7B and Qwen3-ASR-0.6B recognize singing voice and songs accurately. In addition to achieving strong singing-voice recognition, the Qwen3-ASR family also supports direct transcription of complete songs with background music (BGM), demonstrating robustness to accompaniment and complex musical mixtures.
\end{enumerate}

\subsection{Inference Efficiency}

The speed benchmarks of Qwen3-ASR are conducted in two settings: offline batch inference and online asynchronous inference. The former is evaluated using vLLM’s offline batch generation, while the latter is evaluated with a multi-concurrency request setup based on vLLM Serve, which better reflects inference efficiency in industrial environments. All experiments are run with vLLM v0.14.0, with CUDA Graph enabled and bfloat16 precision for inference. The results in~\Cref{tab:efficiency_all} show that, under different concurrency levels, Qwen3-ASR-0.6B can achieve an average Time-to-First-Token (TTFT) as low as 92ms. It reaches a real-time factor~(RTF) as low as 0.064 and throughput as high as 2000 at a concurrency of 128, which means it can process 2,000 seconds of audio per second.

\begin{table}[ht]
\centering
\caption{\textbf{Efficiency of Qwen3-ASR family models. Qwen3-ASR-0.6B and Qwen3-ASR-1.7B support vLLM-based inference in both offline batch and online asynchronous mode, while Qwen3-ForcedAligner-0.6B supports offline batch inference in PyTorch only. All measurements in the table are based on input audio of approximately 2 minutes for ASR and 1 minute for FA in length, and all inference is performed on a single typical computing resource. Conc. denotes the concurrency level. TTFT p95 denotes the 95th percentile TTFT latency.}}
\label{tab:efficiency_all}

\small
\setlength{\tabcolsep}{4pt}
\resizebox{\textwidth}{!}{%
\begin{tabular}{lccccccc}
\toprule
\multirow{2}{*}{\textbf{Model}}& & \multicolumn{2}{c}{\textbf{Offline}} & \multicolumn{4}{c}{\textbf{Online async}} \\
\cmidrule(lr){3-4}\cmidrule(lr){5-8}
 & \textbf{Conc.} &
\textbf{RTF} & \textbf{Throughput} &
\textbf{TTFT avg. (ms)} & \textbf{TTFT p95 (ms)} &
\textbf{RTF} & \textbf{Throughput} \\
\midrule

\multirow{10}{*}{\textbf{Qwen3-ASR-0.6B}}
 & 1   & 0.00923 & 108.34  & 92   & 105  & 0.00940 & 106.38  \\
 & 2   & 0.01124 & 177.94  & 103  & 168  & 0.01108 & 180.51  \\
 & 4   & 0.01284 & 311.53  & 132  & 203  & 0.01224 & 326.80  \\
 & 8   & 0.01600 & 500.00  & 228  & 417  & 0.01472 & 543.48  \\
 & 16  & 0.02384 & 671.14  & 459  & 882  & 0.01936 & 826.45  \\
 & 32  & 0.03808 & 840.34  & 820  & 1575 & 0.02912 & 1098.90 \\
 & 64  & 0.06336 & 1010.10 & 1631 & 3196 & 0.04352 & 1470.59 \\
 & 128 & 0.11264 & 1136.36 & 3210 & 6195 & 0.06400 & 2000.00 \\
 & 256 & 0.21504 & 1190.48 & -   & -   & -      & -      \\
 & 512 & 0.44544 & 1149.43 & -   & -   & -      & -      \\
\midrule

\multirow{10}{*}{\textbf{Qwen3-ASR-1.7B}}
 & 1   & 0.01482 & 67.48   & 102  & 113  & 0.01483 & 67.43   \\
 & 2   & 0.01540 & 129.87  & 117  & 170  & 0.01530 & 130.72  \\
 & 4   & 0.01712 & 233.64  & 135  & 192  & 0.01688 & 236.97  \\
 & 8   & 0.02072 & 386.10  & 224  & 382  & 0.02000 & 400.00  \\
 & 16  & 0.02896 & 552.49  & 443  & 791  & 0.02640 & 606.06  \\
 & 32  & 0.04608 & 694.44  & 847  & 1570 & 0.03968 & 806.45  \\
 & 64  & 0.07360 & 869.57  & 1597 & 2942 & 0.06208 & 1030.93 \\
 & 128 & 0.13056 & 980.39  & 3392 & 6227 & 0.10496 & 1219.51 \\
 & 256 & 0.24320 & 1052.63 & -   & -   & -      & -      \\
 & 512 & 0.50176 & 1020.41 & -   & -   & -      & -      \\
\midrule

\multirow{8}{*}{\textbf{Qwen3-ForcedAligner-0.6B}}
 & 1   & 0.00889 & 112.49  & - & - & - & - \\
 & 2   & 0.00232 & 862.07  & - & - & - & - \\
 & 4   & 0.00432 & 925.93  & - & - & - & - \\
 & 8   & 0.00832 & 961.54  & - & - & - & - \\
 & 16  & 0.01696 & 943.40  & - & - & - & - \\
 & 32  & 0.03584 & 892.86  & - & - & - & - \\
 & 64  & 0.08192 & 781.25  & - & - & - & - \\
 & 128 & 0.19712 & 649.35  & - & - & - & - \\
\bottomrule
\end{tabular}
}
\end{table}

\section{Qwen3-ForcedAligner}
\subsection{Overview}
Qwen3-ForcedAligner-0.6B aims to estimate the start and end timestamps of each word or character in a speech, given the corresponding transcript.
Qwen3-ForcedAligner-0.6B reframes the forced alignment task within a slot-filling formulation. Specifically, given a speech and a transcript augmented with special tokens \textit{[time]} that denote word-level or character-level start and end timestamp slots, Qwen3-ForcedAligner-0.6B directly predicts the corresponding discrete timestamp indices for each slot.

The key features and contributions of Qwen3-ForcedAligner-0.6B can be summarized as:
\begin{itemize}
    \item \textbf{Accurate Timestamp Prediction.} Qwen3-ForcedAligner-0.6B exhibits substantially lower timestamp prediction shifts, achieving a relative reduction of 67\%\textasciitilde77\% in accumulated average shift on the human-labeled test datasets compared with other forced alignment methods.
    \item \textbf{Broad Application Scenarios.} Qwen3-ForcedAligner-0.6B supports speech in 11 languages with durations of up to 300 seconds, including cross-lingual scenarios, and allows users to flexibly customize timestamp prediction for any word or character.
    \item \textbf{Fast Inference Speed.} Qwen3-ForcedAligner-0.6B abandons the next-token prediction paradigm and adopts non-autoregressive~(NAR) inference for timestamp prediction.
\end{itemize}

\subsection{Model Design}

\begin{figure}[h]
    \centering
    \includegraphics[width=0.7\linewidth]{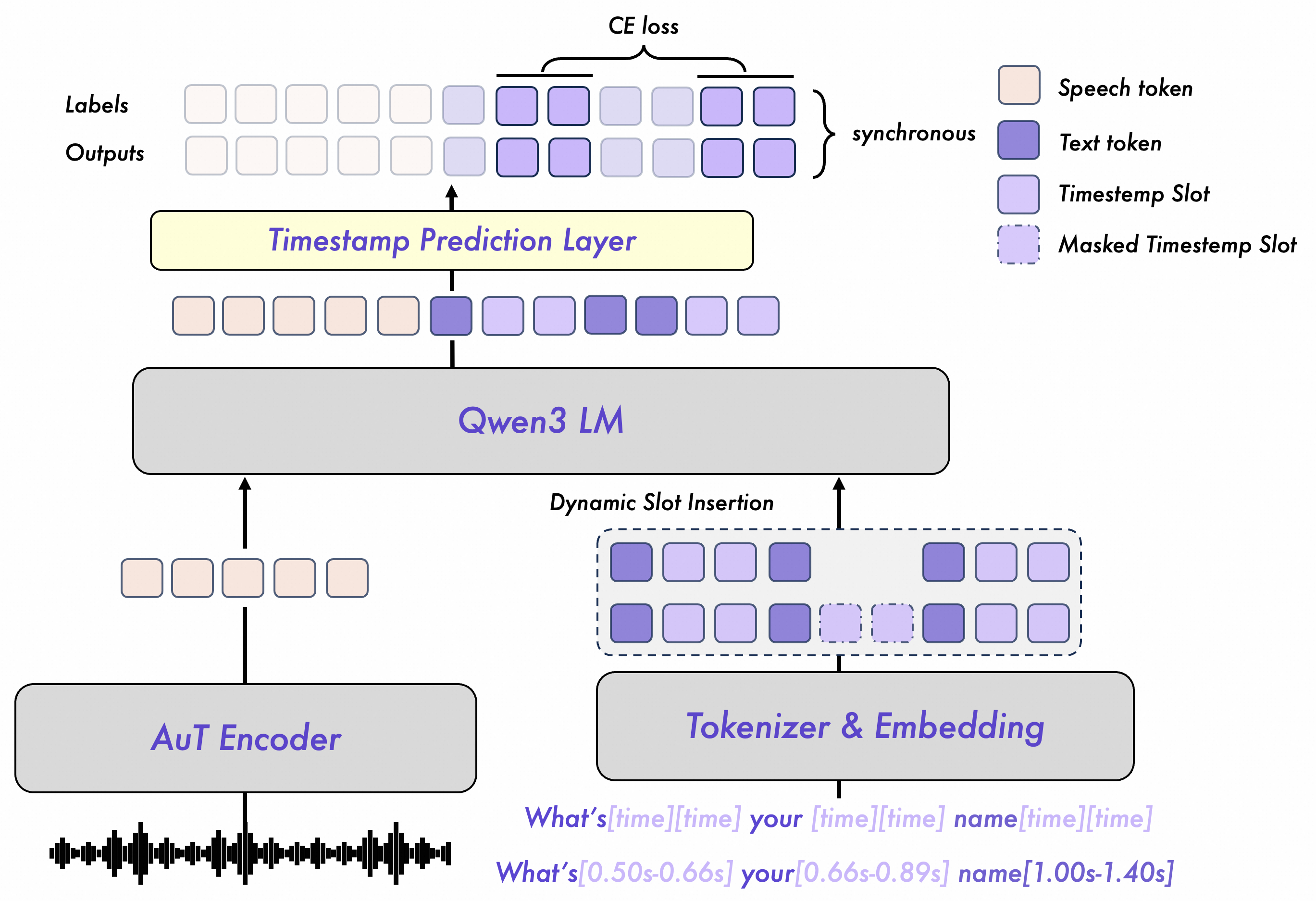}
    \caption{Illustration of Qwen3-ForcedAligner-0.6B. During training, randomly masked timestamp slots are dynamically inserted into the token sequence to represent word or character boundaries. The combined sequence is fed into Qwen3-0.6B LLM, and a timestamp prediction layer predicts the corresponding timestamp indices for each slot. Supervision is applied with cross‑entropy loss on synchronously aligned label and output sequences.}
    \label{fig:fig2}
\end{figure}

As shown in Figure~\ref{fig:fig2}, Qwen3-ForcedAligner-0.6B employs a pretrained AuT encoder to process the input speech signal and obtain speech embeddings.
The transcript is reformatted by appending start and end timestamp labels to each word or character, after which each timestamp label is replaced with a special token \textit{[time]} and fed into the tokenizer.
Moreover, the timestamp labels in the transcript are discretized into indices by dividing each timestamp value by the 80ms frame duration of the AuT encoder output.
Speech and text embedding sequences are processed by the Qwen3-0.6B LLM, followed by a timestamp prediction linear layer that predicts timestamp indices for the entire input sequence. In this work, the maximum number of classes is 3,750, corresponding to support for speech inputs of up to 300s.

The AuT encoder and the multilingual Qwen3-0.6B LLM jointly provide Qwen3-ForcedAligner-0.6B with multilingual and cross-lingual capabilities. Specifically, the AuT encoder, pretrained on a large-scale multilingual corpus, generates effective frame-level speech embeddings for multiple languages, while the multilingual Qwen3-0.6B LLM handles semantic information across different languages. In addition, the special token \textit{[time]} and the timestamp prediction layer do not rely on language-specific phoneme sets or dictionaries. Details can be found in~\cite{llmfa}.
\subsection{Training Strategies}
Training Qwen3-ForcedAligner-0.6B requires word-level or character-level timestamp labels for a large number of speech–transcript pairs. However, because manual annotation is prohibitively expensive, we use pseudo-timestamp labels generated by the Montreal forced aligner (MFA)~\cite{mfa}, which is among the most accurate existing forced alignment methods. It is important to note that MFA pseudo-labels inherently contain noise and systematic shifts. Qwen3‑ForcedAligner does not simply replicate MFA outputs; instead, it distills and smooths these pseudo-labels, resulting in more stable timestamp predictions with reduced shift.

LALMs typically use a training scheme in which the last token of the output sequence and the first token of the label sequence are removed, creating a one-position offset between the two sequences; the cross-entropy loss is then computed to implement the standard next-token prediction paradigm. However, this paradigm is not suitable for filling timestamp slots. Qwen3-ForcedAligner-0.6B employs causal training, keeping the output and label sequences non-shifted, which allows the model to explicitly recognize timestamp slots during training and predict the timestamp indices to fill them. Moreover, causal training enables Qwen3-ForcedAligner-0.6B to incorporate prior contextual information when predicting the timestamp for the current slot, ensuring global consistency in timestamp prediction.
The cross-entropy loss is computed only in the timestamp slots, thereby focusing the training objective of Qwen3-ForcedAligner-0.6B on timestamp slot filling.

In addition, Qwen3-ForcedAligner-0.6B employs a dynamic slot insertion strategy during training to enhance its generalization capability. Specifically, for each word or character in a sample, the model randomly determines whether to insert start and end timestamp slots afterward.
\subsection{Inference and Usability}
Since the token sequences remain non-shifted during training, users can insert start and end timestamp slots after any word or character, and Qwen3-ForcedAligner-0.6B uses non-autoregressive~(NAR) decoding to predict the timestamp indices for all slots in the transcript simultaneously. Once the timestamp indices are obtained, multiplying each index by 80ms recovers the actual predicted timestamps.

The speed benchmark for Qwen3-ForcedAligner is conducted with FlashAttention and bfloat16. Since the model is non-autoregressive, the inference speed difference between Transformers and vLLM is relatively small; therefore, all our benchmarks are run with Transformers. The results in~\Cref{tab:efficiency_all} show that the model can maintain an RTF close to 0.001 even under high concurrency, i.e., it can process 1,000 seconds of audio per second.

%% file: content/experiments.tex
\section{Experiments }

\subsection{Evaluation Details}

\textbf{Baseline Systems}. To validate the Qwen3-ASR family, we conduct comparative evaluations against state-of-the-art (SOTA) closed-source ASR APIs and widely used open-source models. Specifically, we compare Qwen3-ASR with three leading proprietary services: GPT-4o-Transcribe~\citep{gpt4o}, Gemini-2.5-Pro~\citep{gemini2.5}, and Doubao-ASR~\citep{seedasr}. We further include several multilingual open-source baselines, namely Whisper-large-v3~\citep{icml2023-alec-Whisper}, FunASR-MLT-Nano~\citep{funasr-slm}, and GLM-ASR-Nano~\citep{zai-glm-asr-2512}. Together, these baselines represent strong commercial systems and competitive open-source alternatives, enabling a comprehensive evaluation of Qwen3-ASR under representative real-world conditions.

\textbf{Benchmark Introduction}. We adopt a four-part evaluation protocol to measure the speech recognition performance of the proposed Qwen3-ASR series:
\begin{enumerate}[leftmargin=*,itemsep=0.25em]
  \item \textbf{Public benchmarks (English and Chinese).} We evaluate a broad set of public benchmarks~\citep{conneau2023fleurs, ardila2020common, zhang2022wenetspeech, panayotov2015librispeech,  dai2025wenetspeech, li2025wenetspeech} and report the results separately for subsets of English, standard Mandarin and Chinese dialects, including two recently released benchmarks.
  \item \textbf{Internal robustness suite.} We stress-test the model under challenging real-world conditions using a comprehensive in-house suite, covering English speech from multiple countries and accents (16 accent groups in total), 22 Chinese dialect varieties, and difficult scenarios including elderly and children's speech, extremely low signal-to-noise~(SNR) ratios, nonfluent and tongue-twister-like repetitive speech, and multi-speaker Chinese conversational speech. These settings enable a systematic assessment of robustness to accent/dialect variability and complex acoustic and linguistic conditions.
  \item \textbf{Multilingual evaluation.} The model supports ASR for 30 languages. We evaluate on Common Voice, Fleurs, MLS, MLC-SLM~\citep{mlcslm}, and an internally curated test set spanning 15 languages. The language inventory of each benchmark is specified in~\Cref{method:features}. Since Fleurs covers a particularly large and diverse set of languages, we additionally report results on progressively expanded language subsets grouped by language popularity and practical usage for a more fine-grained characterization. Meaning while, we evaluate the language identification performance on the multilingual open-source benchmarks. 
  \item \textbf{Singing voice recognition.} We evaluate singing voice transcription on both public benchmarks and an internal test set. In the internal evaluation, we emphasize long-form transcription where an entire song is provided as a single input, to assess robustness to long-duration audio as well as the distinctive acoustic and rhythmic properties of singing.
\end{enumerate}

\textbf{Evaluation Metrics}. For recognition accuracy, we report either \textbf{word error rate~(WER)} or \textbf{character error rate~(CER)} depending on the language. We use CER for character-based languages~(e.g., Mandarin Chinese, Cantonese, and Korean) and WER for word-delimited languages~(e.g., English, German, and French). When aggregated results are needed (e.g., average performance across multiple languages or dialects), we report the macro-average (i.e., the unweighted mean across languages/dialects). The best result in each table is highlighted in \textbf{bold}. In addition, when Qwen3-ASR-0.6B is the best-performing model after excluding the larger Qwen3-ASR-1.7B, we also highlight it in bold.

For language identification, we report \textbf{language identification accuracy}.

For timestamp accuracy, Qwen3-ForcedAligner uses \textbf{Accumulated Average Shift} (AAS~\cite{shi2023achieving}), where lower values indicate more accurate timestamp predictions. AAS is defined as the mean absolute difference between predicted timestamps and reference timestamps over all timestamp slots in the evaluated datasets:
\begin{equation}
    \mathrm{AAS}=\frac{1}{N}\sum_{i=1}^{N}\left|\hat{n}_i-n_i\right|,
\end{equation}
where $N$ is the total number of timestamp slots, $\hat{n}_i$ denotes the timestamp predicted by Qwen3-ForcedAligner for slot $i$, and $n_i$ is the corresponding reference timestamp obtained from Montreal Forced Aligner (MFA) or manual annotations.

\subsection{English \& Chinese ASR Performance}
\subsubsection{Opensource ASR Benchmarks}

As shown in~\Cref{tab:asr-zh-en-opensource}, Qwen3-ASR delivers consistently strong performance across English, Mandarin Chinese, and multiple Chinese dialect benchmarks. It is competitive with leading commercial APIs while substantially outperforming widely used open-source baselines. Scaling from Qwen3-ASR-0.6B to Qwen3-ASR-1.7B yields clear and stable gains, indicating that the model benefits effectively from increased capacity.

On \textbf{English} benchmarks, Qwen3-ASR performs particularly well on diverse, real-world data (e.g., crowd-sourced or web-collected speech), where distribution shift is more pronounced than in read-speech settings. In these cases, Qwen3-ASR-1.7B achieves the strongest overall results on several datasets, while remaining close to the best-performing systems on standard academic evaluations such as LibriSpeech. Compared with commercial APIs, whose performance can vary substantially across datasets, Qwen3-ASR shows more consistent accuracy across a broad range of English conditions.

On \textbf{Mandarin Chinese}, Qwen3-ASR demonstrates a clear advantage. It delivers the best overall performance on most Mandarin benchmarks in the table and remains reliable on more challenging large-scale evaluations. Notably, on WenetSpeech, which contains diverse acoustic environments and meeting-style speech, Qwen3-ASR outperforms the available baselines by a large margin.

On \textbf{Chinese dialect} benchmarks, Qwen3-ASR maintains strong accuracy under substantial pronunciation and lexical variation. It consistently ranks among the top systems across Cantonese and other dialect datasets, and performs particularly well on more challenging long-utterance settings, demonstrating robustness beyond short, clean test conditions. While a small number of dialect-specific cases favor specialized commercial APIs, Qwen3-ASR remains highly competitive overall and provides a strong general-purpose solution across dialects without per-dialect customization.

Overall,~\Cref{tab:asr-zh-en-opensource} highlights three key advantages of Qwen3-ASR: (i) strong cross-domain generalization on English benchmarks beyond curated read speech, (ii) state-of-the-art accuracy on Mandarin Chinese across multiple public datasets including large-scale, noisy meeting-style speech, and (iii) robust handling of Chinese dialects, with especially strong performance on Cantonese and long/short dialectal speech. These findings demonstrate that Qwen3-ASR delivers strong, reproducible performance across diverse public benchmarks, while also remaining competitive with top-tier closed-source APIs.

\begin{table}[ht]
\centering
\caption{\textbf{Evaluation on English, Mandarin Chinese, and a range of Chinese dialect benchmarks. For the commercial APIs and the open-source Whisper-large-v3 model, we obtained results by running inference on the test sets ourselves due to the absence of published numbers; for FunASR-MLT-Nano, we report the results from its official technical report. "N/A" denotes that we cannot get a reasonable result by the official API. "–" indicates that the corresponding benchmark result is not reported.}}
\vspace{-0.1in}
\label{tab:asr-zh-en-opensource}
\resizebox{\textwidth}{!}{%
\begin{tabular}{@{}clccccc|ccc@{}}
\toprule
\multicolumn{2}{l}{} &
\begin{tabular}[c]{@{}c@{}}\textbf{GPT-4o}\\ \textbf{-Transcribe}\end{tabular} &
\begin{tabular}[c]{@{}c@{}}\textbf{Gemini-2.5}\\ \textbf{-Pro}\end{tabular} &
\textbf{Doubao-ASR} &
\begin{tabular}[c]{@{}c@{}}\textbf{Whisper}\\ \textbf{-large-v3}\end{tabular} &
\begin{tabular}[c]{@{}c@{}}\textbf{Fun-ASR}\\ \textbf{-MLT-Nano}\end{tabular} &
\begin{tabular}[c]{@{}c@{}}\textbf{Qwen3-ASR}\\ \textbf{-0.6B}\end{tabular} &
\begin{tabular}[c]{@{}c@{}}\textbf{Qwen3-ASR}\\ \textbf{-1.7B}\end{tabular}
\\
\midrule
\multicolumn{9}{l}{\textit{\textbf{English~(en)}}} \\ \midrule
    & LibriSpeech    & \multirow{2}{*}{\textbf{1.39}|3.75} & \multirow{2}{*}{2.89|3.56} & \multirow{2}{*}{2.78|5.70} & \multirow{2}{*}{1.51|3.97} & \multirow{2}{*}{1.68|4.03} & \multirow{2}{*}{2.11|4.55} & \multirow{2}{*}{1.63|\textbf{3.38}}  \\
    & \textit{clean | other}    & & &  &   & & &  \\
    & GigaSpeech             & 25.50 & 9.37 & 9.55  & 9.76   & - & \textbf{8.88}  & \textbf{8.45}  \\
    & CV-en                  & 9.08  & 14.49 & 13.78 & 9.90 & 9.90 & 9.92 & \textbf{7.39} \\
    & Fleurs-en              & \textbf{2.40}  & 2.94 & 6.31 & 4.08 & 5.49 & 4.39 & 3.35 \\
    & MLS-en                 & 5.12  & \textbf{3.68} & 7.09   & 4.87 & - & 6.00 & 4.58 \\

    & Tedlium                & 7.69  & 6.15 & 4.91    & 6.84  & - & \textbf{3.85}   & \textbf{4.50} \\
    & VoxPopuli             & 10.29 & 11.36 & 12.12   & 12.05 & - & \textbf{9.96}   & \textbf{9.15}  \\

\midrule
\multicolumn{9}{l}{\textit{\textbf{Chinese~(zh)}}} \\ \midrule
    & WenetSpeech  & \multirow{2}{*}{15.30|32.27} & \multirow{2}{*}{14.43|13.47} & \multirow{2}{*}{N/A}  & \multirow{2}{*}{9.86|19.11} & \multirow{2}{*}{6.35|-} & \multirow{2}{*}{\textbf{5.97}|\textbf{6.88}} & \multirow{2}{*}{\textbf{4.97}|\textbf{5.88}} \\
    & \textit{net | meeting} &    &       &      &         &        & & \\
    & AISHELL-2-test           & 4.24  & 11.62 & 2.85  & 5.06 & - & 3.15 & \textbf{2.71} \\
    & SpeechIO     & 12.86 & 5.30 & 2.93   & 7.56 & - & 3.44 & \textbf{2.88} \\
    & Fleurs-zh           & 2.44  & 2.71 & 2.69 & 4.09 & 3.51 & 2.88 & \textbf{2.41} \\
    & CV-zh     & 6.32  & 7.70 & 5.95    & 12.91 & 6.20 & 6.89 & \textbf{5.35} \\

\midrule
\multicolumn{9}{l}{\textit{\textbf{Chinese Dialect}}} \\ \midrule
    & KeSpeech     & 26.87 & 24.71 & 5.27  & 28.79 & - & 7.08 & \textbf{5.10} \\
    & Fleurs-yue   & 4.98  & 9.43 & 4.98 & 9.18 & - & 5.79 & \textbf{3.98} \\
    & CV-yue       & 11.36 & 18.76 & 13.20  & 16.23 & - & 9.50 & \textbf{7.57} \\
    & CV-zh-tw     & 6.32  & 7.31 & 4.06   & 7.84 & - & 5.59 & \textbf{3.77} \\
    & WenetSpeech-Yue        & \multirow{2}{*}{15.62|25.29}  &   \multirow{2}{*}{25.19|11.23}  & \multirow{2}{*}{9.74|11.40} & \multirow{2}{*}{32.26|46.64} & \multirow{2}{*}{-|-} & \multirow{2}{*}{7.54|9.92} & \multirow{2}{*}{\textbf{5.82}|\textbf{8.85}} \\
    & \textit{short | long}   &  &   &  &      & & & \\
    & WenetSpeech-Chuan      & \multirow{2}{*}{34.81|53.98}   & \multirow{2}{*}{43.79|67.30} & \multirow{2}{*}{\textbf{11.40}|\textbf{20.20}}   &  \multirow{2}{*}{14.35|26.80} & \multirow{2}{*}{-|-} & \multirow{2}{*}{13.92|24.45} & \multirow{2}{*}{11.99|21.63} \\
    & \textit{easy | hard }  &  &   &       &      &  & & \\
\bottomrule
\end{tabular}%
}
\end{table}

\subsubsection{Internal ASR Benchmarks}

\begin{table}[ht]
\centering
\caption{\textbf{Evaluation on internal English and Chinese test sets covering multiple accents and dialects, as well as challenging acoustic conditions and difficult speaking scenarios.}}
\vspace{-0.1in}
\label{tab:asr_internal_zh_en}
\begin{threeparttable}
\resizebox{\textwidth}{!}{%
\begin{tabular}{@{}clccccc|cc@{}}
\toprule
\multicolumn{2}{l}{} &
\begin{tabular}[c]{@{}c@{}}\textbf{GPT-4o}\\ \textbf{-Transcribe}\end{tabular} &
\begin{tabular}[c]{@{}c@{}}\textbf{Gemini-2.5}\\ \textbf{-Pro}\end{tabular} &
\textbf{Doubao-ASR} &
\begin{tabular}[c]{@{}c@{}}\textbf{Whisper}\\ \textbf{-large-v3}\end{tabular} &
\begin{tabular}[c]{@{}c@{}}\textbf{Fun-ASR}\\ \textbf{-MLT-Nano}\end{tabular} &
\begin{tabular}[c]{@{}c@{}}\textbf{Qwen3-ASR}\\ \textbf{-0.6B}\end{tabular} &
\begin{tabular}[c]{@{}c@{}}\textbf{Qwen3-ASR}\\ \textbf{-1.7B}\end{tabular} \\
\midrule
\multicolumn{9}{l}{\textit{\textbf{Accented English}}} \\ \midrule
& Dialog-Accented English & 28.56 & 23.85 & 20.41 & 21.30 & 19.96 & \textbf{16.62} & \textbf{16.07} \\
\midrule
\multicolumn{9}{l}{\textit{\textbf{Chinese Mandarin}}} \\ \midrule
& Elders\&Kids     & 14.27 & 36.93 & 4.17  & 10.61 & 4.54  & 4.48 & \textbf{3.81} \\
& ExtremeNoise     & 36.11 & 29.06 & 17.04 & 63.17 & 36.55 & 17.88 & \textbf{16.17} \\
& TongueTwister    & 20.87 & 4.97  & 3.47  & 16.63 & 9.02  & 4.06 & \textbf{2.44} \\
& Dialog-Mandarin & 20.73 & 12.50 & 6.61  & 14.01 & 7.32  & 7.06 & \textbf{6.54} \\
\midrule
\multicolumn{9}{l}{\textit{\textbf{Chinese Dialect}}} \\ \midrule
& Dialog-Cantonese             & 16.05 & 14.98 & 7.56  & 31.04 & 5.85  & \textbf{4.80} & \textbf{4.12} \\
& Dialog-Chinese Dialects  & 45.37 & 47.70 & 19.85 & 44.55 & 19.41 & \textbf{18.24} & \textbf{15.94} \\
\bottomrule
\end{tabular}%
}
\begin{tablenotes}[flushleft]
\footnotesize
\begin{minipage}{\textwidth}
\item \textbf{Dialect coverage:} Results for \textit{Dialog-Accented English} are averaged over 16 accents, and results for \textit{Dialog-Chinese Dialects} are averaged over 22 Chinese dialects. Detailed category definitions are provided in~\Cref{method:features}.
\end{minipage}
\end{tablenotes}

\end{threeparttable}
\end{table}

To further assess robustness in realistic deployment settings, we evaluate Qwen3-ASR on our internal robustness suite; results are summarized in Table~\ref{tab:asr_internal_zh_en}. Qwen3-ASR delivers consistently strong performance across all subsets, and scaling from 0.6B to 1.7B yields stable gains. In the accented-English evaluation, Qwen3-ASR achieves the lowest WER among all compared systems, surpassing both commercial APIs and open-source baselines, indicating better generalization to accent variation. On Mandarin, Qwen3-ASR-1.7B performs best across all evaluated subsets, demonstrating robustness under difficult acoustic and speaking conditions. In dialectal Chinese, Qwen3-ASR again achieves the best results on both conversational Cantonese and the aggregated 22-dialect evaluation; the gains are particularly pronounced in the multi-dialect mixture, highlighting improved robustness as linguistic diversity increases. Overall, these internal results are consistent with the public-benchmark findings and further confirm that Qwen3-ASR provides reliable recognition quality in high-variability scenarios.

\subsection{ Multilingual ASR and Language Identification}
\subsubsection{Multilingual ASR Performance}

\begin{table}[ht]
\centering
\caption{\textbf{Evaluation of multilingual ASR systems on a comprehensive set of benchmark datasets. }}
\vspace{-0.1in}
\label{tab:asr-multilingual-asr}
\begin{threeparttable}
\resizebox{0.85\textwidth}{!}{%
\begin{tabular}{@{}clccccc@{}}
\toprule
\multicolumn{2}{l}{} &
\begin{tabular}[c]{@{}c@{}}\textbf{GLM-ASR}\\ \textbf{-Nano-2512}\end{tabular} &
\begin{tabular}[c]{@{}c@{}}\textbf{Whisper}\\ \textbf{-large-v3}\end{tabular} &
\begin{tabular}[c]{@{}c@{}}\textbf{Fun-ASR}\\ \textbf{-MLT-Nano}\end{tabular} &
\begin{tabular}[c]{@{}c@{}}\textbf{Qwen3-ASR}\\ \textbf{-0.6B}\end{tabular} &
\begin{tabular}[c]{@{}c@{}}\textbf{Qwen3-ASR}\\ \textbf{-1.7B}\end{tabular} \\
\midrule
\multicolumn{7}{l}{\textit{\textbf{Open-sourced Benchmarks}}} \\
\midrule
& MLS                 & 13.32 & 8.62  & 28.70  & 13.19 & \textbf{8.55}  \\
& CommonVoice         & 19.40 & 10.77 & 17.25  & 12.75 & \textbf{9.18}  \\
& MLC-SLM             & 34.93 & 15.68 & 29.94  & 15.84 & \textbf{12.74}  \\
\midrule
& Fleurs              & 16.08 & 5.27  & 10.03  & 7.57  & \textbf{4.90}  \\
& Fleurs$^{\dagger}$  & 20.05 & 6.85  & 31.89  & 10.37 & \textbf{6.62}  \\
& Fleurs$^{\dagger}$$^{\dagger}$ & 24.83 & \textbf{8.16} & 47.84 & 21.80 & 12.60 \\
\midrule
\multicolumn{7}{l}{\textit{\textbf{Qwen-ASR Internal Benchmarks}}} \\ \midrule
& News-Multilingual   & 49.40 & 14.80 & 65.07  & 17.39 & \textbf{12.80} \\
\bottomrule
\end{tabular}

}
\begin{tablenotes}
\footnotesize
\begin{minipage}{\textwidth}
\item \textbf{Language coverage:} 
\textit{MLS} includes 8 languages: \{da, de, en, es, fr, it, pl, pt\}. \\
\textit{CommonVoice} includes 13 languages: \{en, zh, yue, zh\_TW, ar, de, es, fr, it, ja, ko, pt, ru\}. \\
\textit{MLC-SLM} includes 11 languages: \{en, fr, de, it, pt, es, ja, ko, ru, th, vi\}. \\
\textit{Fleurs} includes 12 languages: \{en, zh, yue, ar, de, es, fr, it, ja, ko, pt, ru \}. \\
\textit{Fleurs$^{\dagger}$} includes 8 additional languages beyond \textit{Fleurs}: \{hi, id, ms, nl, pl, th, tr, vi\}. \\
\textit{Fleurs$^{\dagger\dagger}$} includes 10 additional languages beyond \textit{Fleurs$^{\dagger}$}: \{cs, da, el, fa, fi, fil, hu, mk, ro, sv\}. \\
\textit{News-Multilingual} includes 15 languages: \{ar, de, es, fr, hi, id, it, ja, ko, nl, pl, pt, ru, th, vi\}.
\end{minipage}
\end{tablenotes}
\end{threeparttable}
\end{table}

In this part, we illustrate the multilingual ASR performance of the Qwen3-ASR series on a broad set of public benchmarks as well as our internal multilingual news evaluation (Table~\ref{tab:asr-multilingual-asr}). Overall, Qwen3-ASR-1.7B achieves the best average performance on most test settings, showing strong generalization across languages and domains, while Qwen3-ASR-0.6B provides a competitive lightweight alternative.

On MLS, Common Voice and MLC-SLM benchmarks, Qwen3-ASR-1.7B consistently outperforms the evaluated open-source baselines, including the widely used Whisper-large-v3, and substantially surpasses smaller multilingual models. For Fleurs, which spans more languages and diverse recording conditions, Qwen3-ASR-1.7B achieves the best performance on the 12- and 20-language subsets. However, relative to Whisper-large-v3, its performance degrades on the full 30-language setting, indicating room for improvement in handling increased linguistic diversity and long-tail languages. Nevertheless, Qwen3-ASR-1.7B remains markedly better than the 0.6B variant, suggesting that model scaling improves robustness in more challenging multilingual regimes.

Finally, on our internal News-Multilingual benchmark, Qwen3-ASR-1.7B achieves the best overall performance, demonstrating stronger robustness to domain shift (e.g., broadcast/news-style speech) than all baselines. Overall, these results indicate effective scaling behavior and strong multilingual recognition across both public and internal evaluations. Per-language results for the Qwen3-ASR family are provided in the Appendix.

\subsubsection{Language Identification Performance}

\begin{table}[h]
\centering
\caption{\textbf{Language identification accuracy~(\%) $\uparrow$ on open-source multilingual test sets.}}
\vspace{-0.1in}
\label{tab:langid_fleurs_cv}

\begin{threeparttable}
\resizebox{0.75\textwidth}{!}{%
\begin{tabular}{@{}lccc@{}}
\toprule
& \textbf{Whisper-large-v3} & \textbf{Qwen3-ASR-0.6B} & \textbf{Qwen3-ASR-1.7B} \\ \midrule
MLS           & \textbf{99.9} & 99.3 & \textbf{99.9} \\
CommonVoice   & 92.7 & \textbf{98.2} & \textbf{98.7} \\
MLC-SLM       & 89.2 & \textbf{92.7} & \textbf{94.1} \\
Fleurs        & 94.6 & \textbf{97.1} & \textbf{98.7} \\
\midrule
\textit{\textbf{Avg.}} & 94.1 & \textbf{96.8} & \textbf{97.9} \\
\bottomrule
\end{tabular}%
}

\begin{tablenotes}[flushleft]
\footnotesize
\begin{minipage}{\textwidth}
\item \textbf{Language coverage:} The language sets follow \Cref{tab:asr-multilingual-asr}. Here, Fleurs corresponds to Fleurs$^{\dagger\dagger}$ in \Cref{tab:asr-multilingual-asr} and covers 30 languages.
\end{minipage}
\end{tablenotes}

\end{threeparttable}
\end{table}

Following the output template in~\Cref{section:traning-strategies}, Qwen3-ASR not only decodes speech into text, but also performs language identification (LID) via natural-language prompting before ASR decoding. In this section, we evaluate LID accuracy on 4 multilingual benchmarks: Fleurs (30 languages), MLS~(9 languages),  CommonVoice~(13 languages), MLC-SLM~(11 languages); the covered languages are detailed in~\Cref{method:features}. As shown in Table~\ref{tab:langid_fleurs_cv}, we compare Qwen3-ASR-0.6B and Qwen3-ASR-1.7B with Whisper-large-v3, a strong multilingual ASR model with built-in LID capability. Both Qwen3-ASR models outperform Whisper-large-v3, demonstrating stable and effective language identification across these mainstream languages. Most remaining errors on Fleurs stem from confusion between Malay~(ms) and Indonesian~(id), two closely related languages with high acoustic similarity.

\subsection{Singing Voice \& Songs Recognition Performance}

\begin{table}[ht]
\centering
\caption{\textbf{Singing-voice and song-transcription results. WER~(\%) is reported for singing-only benchmarks and long-form songs with background music. "N/A" indicates that the model does not support long-form song recognition due to the poor performance.}}
\vspace{-0.1in}
\label{tab:singing-and-songs}
\resizebox{\textwidth}{!}{%
\begin{tabular}{@{}clccccc|cc@{}}
\toprule
\multicolumn{2}{l}{} &
\begin{tabular}[c]{@{}c@{}}\textbf{GPT-4o}\\ \textbf{-Transcribe}\end{tabular} &
\begin{tabular}[c]{@{}c@{}}\textbf{Gemini-2.5}\\ \textbf{-Pro}\end{tabular} &
\begin{tabular}[c]{@{}c@{}}\textbf{Doubao-ASR}\\ \textbf{-1.0}\end{tabular} &
\begin{tabular}[c]{@{}c@{}}\textbf{Whisper}\\ \textbf{-large-v3}\end{tabular} &
\begin{tabular}[c]{@{}c@{}}\textbf{Fun-ASR-MLT}\\ \textbf{-Nano}\end{tabular} &
\begin{tabular}[c]{@{}c@{}}\textbf{Qwen3-ASR}\\ \textbf{-1.7B}\end{tabular} &
\\
\midrule
\multicolumn{7}{l}{\textit{\textbf{Singing}}} \\ \midrule
    & M4Singer      & 16.77  & 20.88 & 7.88   & 13.58 & 7.29  & \textbf{5.98}  \\
    & MIR-1k-vocal  & 11.87  & 9.85  & 6.56   & 11.71 & 8.17  & \textbf{6.25} \\   
    & Opencpop      & 7.93   & 6.49  & 3.80   & 9.52  & \textbf{2.98}  & 3.08 \\   
    & Popcs         & 32.84  & 15.13 & 8.97   & 13.77 & 9.42  & \textbf{8.52}  \\   
\midrule
\multicolumn{7}{l}{\textit{\textbf{Songs with BGM}}} \\ \midrule
    & EntireSongs-en & 30.71 & \textbf{12.18} & 33.51 & N/A & N/A & 14.60  \\
    & EntireSongs-zh & 34.86 & 18.68 & 23.99  & N/A & N/A & \textbf{13.91}\\
\bottomrule
\end{tabular}

}
\end{table}

Table~\ref{tab:singing-and-songs} reports results for singing-voice transcription and long-form song transcription with background music. Overall, Qwen3-ASR-1.7B is robust to melody-induced pronunciation variation and musical accompaniment, outperforming most commercial APIs and open-source baselines across the evaluated sets. For \textbf{singing-only} benchmarks, it achieves the best performance for M4Singer, MIR-1k-vocal, and Popcs, while remaining competitive for Opencpop (second to FunASR-MLT-Nano by a small margin), indicating strong generalization across singing styles and recording conditions with reduced sensitivity to pitch drift, phoneme elongation, and rhythmic lyric variation. For \textbf{full songs with background music}, Qwen3-ASR-1.7B substantially outperforms open-source baselines; Whisper-large-v3 and FunASR-MLT-Nano degrade markedly in long-form, music-mixed settings. It achieves high accuracy for both English and Chinese songs, ranking first on the Chinese set and remaining competitive with the best commercial system on the English set, suggesting that Qwen3-ASR is well suited to realistic music-containing scenarios and background-music-robust and narrows the gap between speech ASR and singing/song transcription.

\subsection{Streaming Speech Recognition}

This section evaluates Qwen3-ASR-1.7B and Qwen3-ASR-0.6B in both offline and streaming inference modes. Benefiting from the dynamic attention-window mechanism, the Qwen3-ASR family supports streaming inference naturally. Table~\ref{tab:streaming} reports results on three open-source test sets using a 2-second chunk size, a 5-token fallback, and keeping the last four chunks unfixed. Overall, Qwen3-ASR provides a unified model for offline and streaming use, while streaming inference preserves strong recognition accuracy.

\begin{table}[ht]
\centering
\caption{\textbf{ASR performance of the two inference modes on three open-source benchmarks.}}
\label{tab:streaming}
\resizebox{0.68\textwidth}{!}{%
\begin{tabular}{cccccc}
\toprule
\textbf{Model}        &  \textbf{Infer. Mode}   & \textbf{Librispeech} & \textbf{Fleurs-en} & \textbf{Fleurs-zh} & \textbf{Avg.} \\ 
\midrule
\multirow{2}{*}{\textbf{Qwen3-ASR-1.7B}} & Offline   & 1.63 | 3.38          & 3.35               & 2.41               & 2.69          \\
 & Streaming & 1.95 | 4.51          & 4.02               & 2.84               & 3.33          \\ 
\midrule
\multirow{2}{*}{ \textbf{Qwen3-ASR-0.6B}} & Offline   & 2.11 | 4.55          & 4.39               & 2.88               & 3.48          \\
& Streaming & 2.54 | 6.27          & 5.38               & 3.40               & 4.40          \\ 
\bottomrule
\end{tabular}
}
\end{table}

\subsection{Precision of Timestamps}
\begin{table}[ht]
\centering
\caption{\textbf{Accumulated Average Shift (AAS, ms) $\downarrow$ of Qwen3-ForcedAligner-0.6B and competing forced-alignment methods on MFA-labeled and human-labeled test sets.}}
\label{tab:fa_results}
\resizebox{0.86\textwidth}{!}{%
\begin{tabular}{lcccc}
\toprule
                       & \textbf{Monotonic-Aligner} & \textbf{NFA}   & \textbf{WhisperX} & \textbf{Qwen3-ForcedAligner-0.6B} \\ \midrule
\multicolumn{5}{l}{\textit{\textbf{MFA-Labeled Raw}}}                               \\ \midrule
Chinese                & 161.1             & 109.8 & -        & \textbf{33.1}       \\
English                & -                 & 107.5 & 92.1     & \textbf{37.5}       \\
French                 & -                 & 100.7 & 145.3    & \textbf{41.7}       \\
German                 & -                 & 122.7 & 165.1    & \textbf{46.5}       \\
Italian                & -                 & 142.7 & 155.5    & \textbf{75.5}       \\
Japanese               & -                 & -     & -        & \textbf{42.4}       \\
Korean                 & -                 & -     & -        & \textbf{37.2}       \\
Portuguese             & -                 & -     & -        & \textbf{38.4}       \\
Russian                & -                 & 200.7 & -        & \textbf{40.2}       \\
Spanish                & -                 & 124.7 & 108.0    & \textbf{36.8}       \\
\textit{\textbf{Avg.}} & 161.1             & 129.8 & 133.2    & \textbf{42.9}       \\ \midrule
\multicolumn{5}{l}{\textit{\textbf{MFA-Labeled Concat-300s}}}                        \\ \midrule
Chinese                & 1742.4            & 235.0 & -        & \textbf{36.5}       \\
English                & -                 & 226.7 & 227.2    & \textbf{58.6}       \\
French                 & -                 & 230.6 & 2052.2   & \textbf{53.4}       \\
German                 & -                 & 220.3 & 993.4    & \textbf{62.4}       \\
Italian                & -                 & 290.5 & 5719.4   & \textbf{81.6}       \\
Japanese               & -                 & -     & -        & \textbf{81.3}       \\
Korean                 & -                 & -     & -        & \textbf{42.2}       \\
Portuguese             & -                 & -     & -        & \textbf{50.0}       \\
Russian                & -                 & 283.3 & -        & \textbf{43.0}       \\
Spanish                & -                 & 240.2 & 4549.9   & \textbf{39.6}       \\
Cross-lingual           & -                 & -     & -        & \textbf{34.2}       \\
\textit{\textbf{Avg.}} & 1742.4            & 246.7 & 2708.4   & \textbf{52.9}       \\ \midrule
\multicolumn{5}{l}{\textit{\textbf{Human-Labeled}}}                                 \\ \midrule
Raw                    & 49.9              & 88.6  & -        & \textbf{27.8}       \\
Raw-Noisy              & 53.3              & 89.5  & -        & \textbf{41.8}       \\
Concat-60s              & 51.1              & 86.7  & -        & \textbf{25.3}       \\
Concat-300s             & 410.8             & 140.0 & -        & \textbf{24.8}       \\
Concat-Cross-lingual     & -                 & -     & -        & \textbf{42.5}       \\
\textit{\textbf{Avg.}} & 141.3             & 101.2 & -        & \textbf{32.4}       \\ \bottomrule
\end{tabular}
}
\end{table}



Table~\ref{tab:fa_results} reports the AAS of Qwen3-ForcedAligner-0.6B and competing forced-alignment methods on MFA-labeled and human-labeled test sets. Competing methods require language-specific models and support only a limited set of languages, whereas Qwen3-ForcedAligner-0.6B covers multiple languages with a single model and supports cross-lingual, code-switched scenarios. In addition, Qwen3-ForcedAligner-0.6B performs consistently on both short and long utterances, while baseline methods show a sharp degradation in timestamp accuracy on long utterances. Although trained with MFA pseudo-labels, Qwen3-ForcedAligner-0.6B still achieves low AAS on the human-labeled test sets, indicating strong real-world generalization.

%% file: content/conclusion.tex
\section{Conclusion}
\label{sec:conclusion}


We present Qwen3-ASR, a model family comprising two automatic speech recognition (ASR) systems and a forced-alignment (FA) model trained on large-scale speech corpora. By leveraging the strong audio understanding capability of the foundation model Qwen3-Omni and a four-stage training pipeline, Qwen3-ASR-1.7B and Qwen3-ASR-0.6B outperform competing models of comparable or larger size, as well as commercial APIs, in both speech coverage and recognition accuracy. The models support language identification and ASR across 30 languages, deliver robust performance in complex acoustic conditions, exhibit resilience to accents and dialects, and maintain effectiveness on singing voice and other real-world speech scenarios. In addition, we introduce Qwen3-ForcedAligner-0.6B, an LLM-based non-autoregressive timestamp predictor that enables forced alignment for 11 languages with end-to-end processing times under five minutes. This approach surpasses three mainstream end-to-end ASR-based FA solutions in timestamp accuracy, inference speed, and language coverage. Alongside releasing the weights for all three models, we open-source a unified and user-friendly inference framework. Overall, the Qwen3-ASR family achieves state-of-the-art performance on real-world evaluations and public benchmarks, and the open-sourced forced-alignment model addresses a critical gap in the speech technology stack. We will continue to advance this open model family in accuracy and functionality.

%% file: content/authors.tex
\section{Authors}
\textbf{Core Contributors:} Xian Shi, Xiong Wang, Zhifang Guo, Yongqi Wang, Pei Zhang, Xinyu Zhang, Zishan Guo, Hongkun Hao, Yu Xi, Baosong Yang, Jin Xu$^{\dag}$, Jingren Zhou, Junyang Lin$^{\dag}$

\textbf{Contributors}\footnote{Alphabetical order. $^{\dag}$Corresponding Authors.}: Yunfei Chu, Daren Chen, Ting He, Hangrui Hu, Jiayi Leng, Zheng Li, Yuanjun Lv, Bingshen Mu, Hao Su, Xian Yang, Xuechun Wang, Yuezhang Wang, Zhenglin Wang, Lei Xie, Jianwei Zhang, Xinfa Zhu, Guangdong Zhou


%% file: content/appendix.tex
\newpage
\appendix

\counterwithin{table}{section}
\renewcommand{\thetable}{\Alph{section}.\arabic{table}}

\section*{Appendix}              
\setcounter{section}{1}          
\setcounter{table}{0}

\begin{table}[ht]
\centering
\caption{\textbf{Evaluation on English, Chinese and a range of Chinese dialect benchmarks. As a member of Qwen3-ASR family, Qwen3-ASR-Flash-1208 serves as an API and its results are for reference in the table.}}
\vspace{-0.1in}
\label{tab:asr-zh-en-opensource2}
\resizebox{0.85\textwidth}{!}{%
\begin{tabular}{@{}clccc@{}}
\toprule
\multicolumn{2}{l}{} &
\begin{tabular}[c]{@{}c@{}}\textbf{Qwen3-ASR-0.6B}\end{tabular} &
\begin{tabular}[c]{@{}c@{}}\textbf{Qwen3-ASR-1.7B}\end{tabular} &
\begin{tabular}[c]{@{}c@{}}\textbf{Qwen3-ASR-Flash-1208}\end{tabular}
\\
\midrule
\multicolumn{5}{l}{\textit{\textbf{English~(en)}}} \\ \midrule
    & LibriSpeech    & \multirow{2}{*}{2.11|4.55} & \multirow{2}{*}{1.63|3.38} & \multirow{2}{*}{1.33|2.40} \\
    & \textit{clean | other}    & & &  \\
    & GigaSpeech             & 8.88  & 8.45  & 8.82 \\
    & CV-en                  & 9.92  & 7.39  & 6.06 \\
    & Fleurs-en              & 4.39  & 3.35  & 2.72 \\
    & MLS-en                 & 6.00  & 4.58  & 3.63 \\
    & Tedlium                & 3.85  & 4.50  & 4.84 \\
    & VoxPopuli              & 9.96  & 9.15  & 8.45 \\
\midrule
\multicolumn{5}{l}{\textit{\textbf{Chinese~(zh)}}} \\ \midrule
    & WenetSpeech  & \multirow{2}{*}{5.97|6.88} & \multirow{2}{*}{4.97|5.88} & \multirow{2}{*}{4.60|5.80} \\
    & \textit{net | meeting} & & & \\
    & AISHELL-2-test         & 3.15  & 2.71  & 2.53 \\
    & SpeechIO               & 3.44  & 2.88  & 2.62 \\
    & Fleurs-zh              & 2.88  & 2.41  & 2.38 \\
    & CV-zh                  & 6.89  & 5.35  & 4.45 \\
\midrule
\multicolumn{5}{l}{\textit{\textbf{Chinese Dialect}}} \\ \midrule
    & KeSpeech               & 7.08  & 5.10  & 3.28 \\
    & Fleurs-yue             & 5.79  & 3.98  & 3.50 \\
    & CV-yue                 & 9.50  & 7.57  & 4.86 \\
    & CV-zh-tw               & 5.59  & 3.77  & 3.30 \\
    & WenetSpeech-Yue        & \multirow{2}{*}{7.54|9.92} & \multirow{2}{*}{5.82|8.85} & \multirow{2}{*}{5.84|8.20} \\
    & \textit{short | long}  &  &  &  \\
    & WenetSpeech-Chuan      & \multirow{2}{*}{13.92|24.45} & \multirow{2}{*}{11.99|21.63} & \multirow{2}{*}{11.52|20.82} \\
    & \textit{easy | hard }  &  &  &  \\
\bottomrule
\end{tabular}%
}
\end{table}

\begin{table}[ht]
\centering
\caption{\textbf{Evaluation of Qwen3-ASR on open-source multilingual benchmarks. As a member of Qwen3-ASR family, Qwen3-ASR-Flash-1208 serves as an API and its results are for reference in the table.}}
\label{tab:qwen_asr_multilingual_side_by_side}

\begin{subtable}[t]{0.49\textwidth}
\centering
\caption{\textbf{MLS, CommonVoice, and MLC-SLM.}}
\label{tab:qwen_asr_multilingual_complete_results}
\resizebox{\linewidth}{!}{%
\begin{tabular}{@{}clcc|c@{}}
\toprule
\multicolumn{2}{l}{} &
\begin{tabular}[c]{@{}c@{}}\textbf{Qwen3-ASR}\\ \textbf{-0.6B}\end{tabular} &
\begin{tabular}[c]{@{}c@{}}\textbf{Qwen3-ASR}\\ \textbf{-1.7B}\end{tabular} &
\begin{tabular}[c]{@{}c@{}}{\textbf{Qwen3-ASR}}\\ {\textbf{-Flash-1208}}\end{tabular} \\
\midrule
\multicolumn{5}{l}{\textit{\textbf{MLS}}} \\
\midrule
&  da  & 16.79 & 11.73 & 7.58 \\
&  de  & 9.52  & 6.05  & 4.11 \\
&  en  & 6.04  & 4.58  & 3.63 \\
&  es  & 7.19  & 4.63  & 3.29 \\
&  fr  & 8.55  & 5.26  & 3.16 \\
&  it  & 19.21 & 13.20 & 7.88 \\
&  pl  & 26.09 & 15.26 & 9.76 \\
&  pt  & 12.16 & 7.71  & 6.83 \\
\midrule
\multicolumn{5}{l}{\textit{\textbf{CommonVoice}}} \\
\midrule
&  ar     & 45.99 & 37.97 & 33.86 \\
&  de     & 9.44  & 5.85  & 3.53  \\
&  en     & 9.92  & 7.39  & 6.06  \\
&  es     & 7.16  & 4.65  & 3.14  \\
&  fr     & 12.25 & 8.56  & 5.88  \\
&  it     & 10.16 & 5.40  & 3.21  \\
&  ja     & 14.96 & 11.64 & 9.31  \\
&  ko     & 8.48  & 5.88  & 3.82  \\
&  pt     & 11.30 & 7.10  & 5.42  \\
&  ru     & 14.07 & 8.28  & 5.73  \\
&  yue    & 9.50  & 7.57  & 4.86  \\
&  zh     & 6.89  & 5.35  & 4.45  \\
&  zh\_tw & 5.59  & 3.77  & 3.30  \\
\midrule
\multicolumn{5}{l}{\textit{\textbf{MLC-SLM}}} \\
\midrule
&  de & 19.78 & 17.19 & 15.76 \\
&  en & 7.44  & 6.41  & 6.55  \\
&  es & 13.89 & 11.07 & 9.31  \\
&  fr & 22.96 & 20.75 & 22.98 \\
&  it & 21.31 & 16.75 & 14.93 \\
&  ja & 14.74 & 11.80 & 9.74  \\
&  ko & 10.31 & 8.61  & 8.09  \\
&  pt & 34.97 & 26.64 & 28.14 \\
&  ru & 19.24 & 15.17 & 13.16 \\
&  th & 19.51 & 14.34 & 19.66 \\
&  vi & 17.67 & 14.92 & 13.11 \\

\bottomrule
\end{tabular}%
}
\end{subtable}\hfill
\begin{subtable}[t]{0.49\textwidth}
\centering
\caption{\textbf{Fleurs.}}
\label{tab:qwen_asr_multilingual_complete_results_fleurs}
\resizebox{\linewidth}{!}{%
\begin{tabular}{@{}clcc|c@{}}
\toprule
\multicolumn{2}{l}{} &
\begin{tabular}[c]{@{}c@{}}\textbf{Qwen3-ASR}\\ \textbf{-0.6B}\end{tabular} &
\begin{tabular}[c]{@{}c@{}}\textbf{Qwen3-ASR}\\ \textbf{-1.7B}\end{tabular} &
\begin{tabular}[c]{@{}c@{}}{\textbf{Qwen3-ASR}}\\ {\textbf{-Flash-1208}}\end{tabular} \\
\midrule
\multicolumn{5}{l}{\textit{\textbf{Fleurs}}} \\
\midrule
& ar  & 25.51 & 16.98 & 14.78 \\
& cs  & 47.67 & 22.42 & 18.68 \\
& da  & 36.36 & 21.00 & 11.85 \\
& de  & 6.48  & 3.92  & 3.03  \\
& el  & 49.67 & 28.08 & 13.85 \\
& en  & 4.39  & 3.35  & 2.72  \\
& es  & 4.94  & 3.36  & 2.68  \\
& fa  & 53.76 & 29.90 & 18.37 \\
& fi  & 46.59 & 25.23 & 12.21 \\
& fil & 36.10 & 24.29 & 19.17 \\
& fr  & 7.72  & 4.75  & 3.44  \\
& hi  & 19.12 & 17.15 & 13.77 \\
& hu  & 59.47 & 34.22 & 21.77 \\
& id  & 7.92  & 5.16  & 3.65  \\
& it  & 4.99  & 2.41  & 1.60  \\
& ja  & 8.33  & 5.20  & 3.09  \\
& ko  & 3.72  & 2.57  & 2.07  \\
& mk  & 37.26 & 19.05 & --    \\
& ms  & 17.66 & 10.39 & 11.37 \\
& nl  & 14.02 & 7.04  & 4.35  \\
& pl  & 24.71 & 12.54 & 7.24  \\
& pt  & 6.21  & 3.92  & 3.18  \\
& ro  & 44.26 & 20.70 & 10.45 \\
& ru  & 9.91  & 5.99  & 4.81  \\
& sv  & 35.87 & 19.36 & 15.02 \\
& th  & 8.34  & 6.32  & 5.53  \\
& tr  & 16.18 & 9.47  & 6.13  \\
& vi  & 8.52  & 5.55  & 3.64  \\
& yue & 5.79  & 3.98  & 3.50  \\
& zh  & 2.88  & 2.41  & 2.38  \\
\bottomrule
\end{tabular}%
}
\end{subtable}

\end{table}

%% file: biblio.bib
@misc{funasr-slm,
      title={{Fun-ASR Technical Report}},
      author={Keyu An and Yanni Chen and Zhigao Chen and Chong Deng and Zhihao Du and Changfeng Gao and Zhifu Gao and Bo Gong and Xiangang Li and Yabin Li and Ying Liu and Xiang Lv and Yunjie Ji and Yiheng Jiang and Bin Ma and Haoneng Luo and Chongjia Ni and Zexu Pan and Yiping Peng and Zhendong Peng and Peiyao Wang and Hao Wang and Haoxu Wang and Wen Wang and Wupeng Wang and Yuzhong Wu and Biao Tian and Zhentao Tan and Nan Yang and Bin Yuan and Jieping Ye and Jixing Yu and Qinglin Zhang and Kun Zou and Han Zhao and Shengkui Zhao and Jingren Zhou and Yanqiao Zhu},
      year={2025},
      url={https://arxiv.org/abs/2509.12508},
}

@misc{seedasr,
  author       = {Ye Bai and
                  Jingping Chen and
                  Jitong Chen and
                  Wei Chen and
                  Zhuo Chen and
                  Chuang Ding and
                  Linhao Dong and
                  Qianqian Dong and
                  Yujiao Du and
                  Kepan Gao and
                  Lu Gao and
                  Yi Guo and
                  Minglun Han and
                  Ting Han and
                  Wenchao Hu and
                  Xinying Hu and
                  Yuxiang Hu and
                  Deyu Hua and
                  Lu Huang and
                  Mingkun Huang and
                  Youjia Huang and
                  Jishuo Jin and
                  Fanliu Kong and
                  Zongwei Lan and
                  Tianyu Li and
                  Xiaoyang Li and
                  Zeyang Li and
                  Zehua Lin and
                  Rui Liu and
                  Shouda Liu and
                  Lu Lu and
                  Yizhou Lu and
                  Jingting Ma and
                  Shengtao Ma and
                  Yulin Pei and
                  Chen Shen and
                  Tian Tan and
                  Xiaogang Tian and
                  Ming Tu and
                  Bo Wang and
                  Hao Wang and
                  Yuping Wang and
                  Yuxuan Wang and
                  Hanzhang Xia and
                  Rui Xia and
                  Shuangyi Xie and
                  Hongmin Xu and
                  Meng Yang and
                  Bihong Zhang and
                  Jun Zhang and
                  Wanyi Zhang and
                  Yang Zhang and
                  Yawei Zhang and
                  Yijie Zheng and
                  Ming Zou},
  title        = {{Seed-ASR: Understanding Diverse Speech and Contexts with LLM-based
                  Speech Recognition}},
  year={2024},
  url          = {https://arxiv.org/abs/2407.04675},
}

@inproceedings{icassp2016-William-las,
  author       = {William Chan and
                  Navdeep Jaitly and
                  Quoc V. Le and
                  Oriol Vinyals},
  title        = {{Listen, attend and spell: {A} neural network for large vocabulary
                  conversational speech recognition}},
  booktitle    = {Proc. ICASSP},
  pages        = {4960--4964},
  publisher    = {{IEEE}},
  year         = {2016},
  url          = {https://doi.org/10.1109/ICASSP.2016.7472621},
}

@misc{gemini2.5,
  author       = {Comanici, Gheorghe and Bieber, Eric and Schaekermann, Mike and Pasupat, Ice and Sachdeva, Noveen and Dhillon, Inderjit and Blistein, Marcel and Ram, Ori and Zhang, Dan and Rosen, Evan and others},
  title        = {{Gemini 2.5: Pushing the Frontier with Advanced Reasoning, Multimodality,
                  Long Context, and Next Generation Agentic Capabilities}},
  year         = {2025},
  url          = {https://arxiv.org/abs/2507.06261},
}

@misc{arxiv2012-alex-rnnt,
  author       = {Alex Graves},
  title        = {{Sequence Transduction with Recurrent Neural Networks}},
  year         = {2012},
  url          = {http://arxiv.org/abs/1211.3711},
}

@inproceedings{Ludwig2020ctc,
  author       = {Ludwig K{\"{u}}rzinger and
                  Dominik Winkelbauer and
                  Lujun Li and
                  Tobias Watzel and
                  Gerhard Rigoll},
  title        = {{CTC-Segmentation of Large Corpora for German End-to-End Speech Recognition}},
  booktitle    = {Proc. Speech and Computer},
  url          = {https://doi.org/10.1007/978-3-030-60276-5\_27},
  pages        = {267--278},
  year         = {2020},
}

@inproceedings{mfa,
  author       = {Michael McAuliffe and
                  Michaela Socolof and
                  Sarah Mihuc and
                  Michael Wagner and
                  Morgan Sonderegger},
  title        = {{Montreal Forced Aligner: Trainable Text-Speech Alignment Using Kaldi}},
  booktitle    = {Proc. Interspeech},
  pages        = {498--502},
  year         = {2017},
  url          = {https://doi.org/10.21437/Interspeech.2017-1386},
}

@misc{gpt4o,
  title = {Hello {GPT-4o}},
  author = {{OpenAI}},
  url = {https://openai.com/index/hello-gpt-4o/},
  year = {2024}
}

@inproceedings{icml2023-alec-Whisper,
  author       = {Alec Radford and
                  Jong Wook Kim and
                  Tao Xu and
                  Greg Brockman and
                  Christine McLeavey and
                  Ilya Sutskever},
  title        = {{Robust Speech Recognition via Large-Scale Weak Supervision}},
  booktitle    = {Proc. ICML},
  pages        = {28492--28518},
  year         = {2023},
  url          = {https://proceedings.mlr.press/v202/radford23a.html},
}

@inproceedings{Elena2023nfa,
  author       = {Elena Rastorgueva and
                  Vitaly Lavrukhin and
                  Boris Ginsburg},
  title        = {{NeMo Forced Aligner and its application to word alignment for subtitle
                  generation}},
  booktitle    = {Proc. Interspeech},
  pages        = {5257--5258},
  year         = {2023},
  url          = {https://www.isca-archive.org/interspeech\_2023/rastorgueva23\_interspeech.html},
}

@inproceedings{shi2023achieving,   
title={{Achieving timestamp prediction while recognizing with non-autoregressive end-to-end ASR model}},   
author={Shi, Xian and Chen, Yanni and Zhang, Shiliang and Yan, Zhijie},   
booktitle={Proc. NCMMSC},    
year={2023},  
}

@misc{arxiv2025-jinxu-qwen3omni,
  author       = {Jin Xu and
                  Zhifang Guo and
                  Hangrui Hu and
                  Yunfei Chu and
                  Xiong Wang and
                  Jinzheng He and
                  Yuxuan Wang and
                  Xian Shi and
                  Ting He and
                  Xinfa Zhu and
                  Yuanjun Lv and
                  Yongqi Wang and
                  Dake Guo and
                  He Wang and
                  Linhan Ma and
                  Pei Zhang and
                  Xinyu Zhang and
                  Hongkun Hao and
                  Zishan Guo and
                  Baosong Yang and
                  Bin Zhang and
                  Ziyang Ma and
                  Xipin Wei and
                  Shuai Bai and
                  Keqin Chen and
                  Xuejing Liu and
                  Peng Wang and
                  Mingkun Yang and
                  Dayiheng Liu and
                  Xingzhang Ren and
                  Bo Zheng and
                  Rui Men and
                  Fan Zhou and
                  Bowen Yu and
                  Jianxin Yang and
                  Le Yu and
                  Jingren Zhou and
                  Junyang Lin},
  title        = {{Qwen3-Omni Technical Report}},
  year         = {2025},
  url          = {https://arxiv.org/abs/2509.17765},
}

@misc{zai-glm-asr-2512,
  author       = {{Z.ai}},
  title        = {GLM ASR 2512},
  howpublished = {\url{https://docs.z.ai/guides/audio/glm-asr-2512}},
  note         = {Accessed: 2026-01-26},
  year         = {2025},
}

@misc{zheng2025group,
  author       = {Chujie Zheng and
                  Shixuan Liu and
                  Mingze Li and
                  Xiong{-}Hui Chen and
                  Bowen Yu and
                  Chang Gao and
                  Kai Dang and
                  Yuqiong Liu and
                  Rui Men and
                  An Yang and
                  Jingren Zhou and
                  Junyang Lin},
  title        = {{Group Sequence Policy Optimization}},
  year         = {2025},
  url          = {https://arxiv.org/abs/2507.18071},
}

@misc{llmfa,
  author       = {Bingshen Mu and
                  Xian Shi and
                  Xiong Wang and
                  Hexin Liu and
                  Jin Xu and
                  Lei Xie},
  title        = {{LLM-ForcedAligner: A Non-Autoregressive and Accurate LLM-Based Forced Aligner for Multilingual and Long-Form Speech}},
  year         = {2026},
  url          = {https://arxiv.org/abs/2601.18220},
}

@inproceedings{mlcslm,
  author       = {Bingshen Mu and
                  Pengcheng Guo and
                  Zhaokai Sun and
                  Shuai Wang and
                  Hexin Liu and
                  Mingchen Shao and
                  Lei Xie and
                  Eng Siong Chng and
                  Longshuai Xiao and
                  Qiangze Feng and
                  Daliang Wang},
  title        = {{Summary on The Multilingual Conversational Speech Language Model Challenge:
                  Datasets, Tasks, Baselines, and Methods}},
  booktitle    = {Proc. ICASSP},
  year         = {2026},
}

@article{dai2025wenetspeech,
  title={Wenetspeech-chuan: A large-scale sichuanese corpus with rich annotation for dialectal speech processing},
  author={Dai, Yuhang and Zhang, Ziyu and Wang, Shuai and Li, Longhao and Guo, Zhao and Zuo, Tianlun and Wang, Shuiyuan and Xue, Hongfei and Wang, Chengyou and Wang, Qing and others},
  journal={arXiv preprint arXiv:2509.18004},
  year={2025}
}

@article{li2025wenetspeech,
  title={Wenetspeech-yue: A large-scale cantonese speech corpus with multi-dimensional annotation},
  author={Li, Longhao and Guo, Zhao and Chen, Hongjie and Dai, Yuhang and Zhang, Ziyu and Xue, Hongfei and Zuo, Tianlun and Wang, Chengyou and Wang, Shuiyuan and Li, Jie and others},
  journal={arXiv preprint arXiv:2509.03959},
  year={2025}
}

@inproceedings{conneau2023fleurs,
  title={Fleurs: Few-shot learning evaluation of universal representations of speech},
  author={Conneau, Alexis and Ma, Min and Khanuja, Simran and Zhang, Yu and Axelrod, Vera and Dalmia, Siddharth and Riesa, Jason and Rivera, Clara and Bapna, Ankur},
  booktitle={2022 IEEE Spoken Language Technology Workshop (SLT)},
  pages={798--805},
  year={2023},
  organization={IEEE}
}

@inproceedings{ardila2020common,
  title={Common voice: A massively-multilingual speech corpus},
  author={Ardila, Rosana and Branson, Megan and Davis, Kelly and Kohler, Michael and Meyer, Josh and Henretty, Michael and Morais, Reuben and Saunders, Lindsay and Tyers, Francis and Weber, Gregor},
  booktitle={Proceedings of the twelfth language resources and evaluation conference},
  pages={4218--4222},
  year={2020}
}

@inproceedings{zhang2022wenetspeech,
  title={Wenetspeech: A 10000+ hours multi-domain mandarin corpus for speech recognition},
  author={Zhang, Binbin and Lv, Hang and Guo, Pengcheng and Shao, Qijie and Yang, Chao and Xie, Lei and Xu, Xin and Bu, Hui and Chen, Xiaoyu and Zeng, Chenchen and others},
  booktitle={Proc. ICASSP},
  pages={6182--6186},
  year={2022},
}

@inproceedings{panayotov2015librispeech,
  title={Librispeech: an asr corpus based on public domain audio books},
  author={Panayotov, Vassil and Chen, Guoguo and Povey, Daniel and Khudanpur, Sanjeev},
  booktitle={2015 IEEE international conference on acoustics, speech and signal processing (ICASSP)},
  pages={5206--5210},
  year={2015},
  organization={IEEE}
}

@article{qwen3-omni,
  author       = {Jin Xu and
                  Zhifang Guo and
                  Hangrui Hu and
                  Yunfei Chu and
                  Xiong Wang and
                  Jinzheng He and
                  Yuxuan Wang and
                  Xian Shi and
                  Ting He and
                  Xinfa Zhu and
                  Yuanjun Lv and
                  Yongqi Wang and
                  Dake Guo and
                  He Wang and
                  Linhan Ma and
                  Pei Zhang and
                  Xinyu Zhang and
                  Hongkun Hao and
                  Zishan Guo and
                  Baosong Yang and
                  Bin Zhang and
                  Ziyang Ma and
                  Xipin Wei and
                  Shuai Bai and
                  Keqin Chen and
                  Xuejing Liu and
                  Peng Wang and
                  Mingkun Yang and
                  Dayiheng Liu and
                  Xingzhang Ren and
                  Bo Zheng and
                  Rui Men and
                  Fan Zhou and
                  Bowen Yu and
                  Jianxin Yang and
                  Le Yu and
                  Jingren Zhou and
                  Junyang Lin},
  title        = {Qwen3-Omni Technical Report},
  journal      = {CoRR},
  volume       = {abs/2509.17765},
  year         = {2025},
}
